# Micromanipulation System for Microscale Magnetic Component Alignment and Assembly

Oliver J. Shindell[1], Aaron C. Davis[1], and David J. Cappelleri[1,2]

*Abstract*—This paper presents a contact-based micromanipulation system for the alignment and installment of microscale magnets into micro robots and devices. Affixing tweezers to a three degree of freedom micromanipulator allows for precise movement of objects. The use of non-magnetic tweezers permits the assembly of magnetized robots, and a magnetic rotating stage allows multiple magnets to be installed into one device in different orientations. By re-orienting the tweezers on the micromanipulator at defined ninety-degree angles, it is possible to assemble a device with magnets oriented in any direction on XY, XZ, and YZ planes. This system is highly precise and flexible, and can be implemented with minimal custom-made parts, making it ideal for development of new magnetic technologies at the microscale.

## I. Introduction

Advancements in the field of magnetically controlled micro-robots have demonstrated their strong potential for medical applications [1]. This utility is owed to the non-invasive nature of magnetic control [2]. In order to harness the potential of magnetically controlled microrobots, an effective assembly system is indispensable.

There are many methods by which micro-assemblies are currently performed, which vary in complexity [3]. A simple method of assembly is to move pieces by hand using tweezers. However, even for surgeons, hands experience tremors of 0.725 mm amplitude on average, which makes assemblies at the microscale extremely challenging [4]. There are existing solutions to this that can manipulate objects as small as 20 µm [5]. However, many of these systems lack a means to orient magnets, which is vital for assembling robots capable of complex and precise movements [6-7].

The system presented in this paper is designed for the assembly of microrobots or devices containing embedded permanent block magnets, such as those described in [8]. The system can pick and place objects as small as 30µm, and contains a magnetic field generated by a permanent magnet, which both aligns other magnets and holds them in place on a stage [9]. This allows magnets to be inserted into a robot in any orientation, making the system highly effective for the assembly of microrobots involving permanent magnets, as well as other microscale devices.

*This work was supported by the National Science Foundation under NSF IIS Award #: 1763689 and NSF CMMI Award #: 2018570 and the National Institute of Health under NIH Award #: 1U01TR004239-01

[1]Oliver J. Shindell, Aaron C. Davis, and David J. Cappelleri are with the School of Mechanical Engineering, Purdue University, West Lafayette, IN 47907, USA. {davi1381,dcappell}@purdue.edu

[2]David J. Cappelleri is also with the Weldon School of Biomedical Engineering (By Courtesy), Purdue University, West Lafayette, IN 47907, USA.

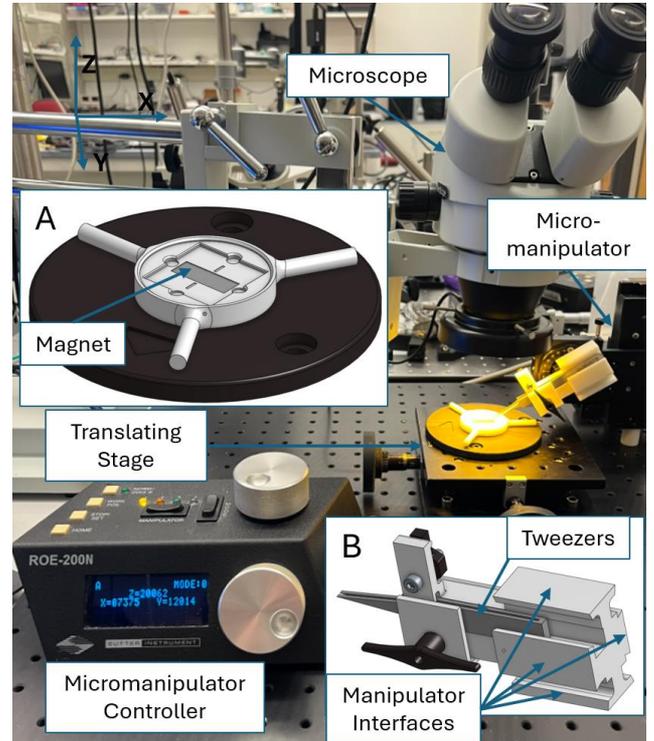

Fig 1: Assembly system. (A). Rotating stage with embedded magnet (B). Tweezer actuation system.

## II. System Overview

The assembly system developed is shown in Fig 1. The system is made up of a combination of custom-made parts, all of which are 3D printed (Formlabs), and off-the-shelf parts. The off-the-shelf parts consist of a micromanipulator (Sutter Instruments), a binocular microscope (AmScope), non-magnetic tweezers (Dumont), a translating XY stage, a neodymium magnet, and an optical breadboard. A yellow light is affixed to the microscope to prevent curing of photosensitive resin.

There are two custom assemblies within the system. The first of these is the rotating stage, seen in Fig 1(A). It is attached to the translating stage to be able to precisely position the assemblies within the microscope field of view. The rotating stage contains a magnet, and by rotating the stage, one can orient a magnetic field in any direction in the XY plane. This allows robots to have magnets poled in different directions, which enables a greater variety of motions.

The second assembly is the tweezer actuation system, seen in Fig 1(B). This system holds one side of the tweezers in place, while the other side is opened or closed with a hand dial that moves a nut along a fine threaded bolt. The tweezers can be removed and replaced if they are damaged. Additionally, the entire tweezer actuation assembly can be attached to the micromanipulator at four angles, while maintaining its grip on an object. This allows objects to be reoriented in YZ and XZ planes, which, when used in conjunction with the rotating stage, allows a magnet to be inserted in any orientation.

A permanent magnet in the base is used because this system is intended to assemble other permanent magnets. To assemble a robot with multiple micromagnets, the attraction force from the base must be greater than the attraction from another micromagnet in the device. This is best accomplished using a permanent magnet of the same material as the micromagnets being assembled.

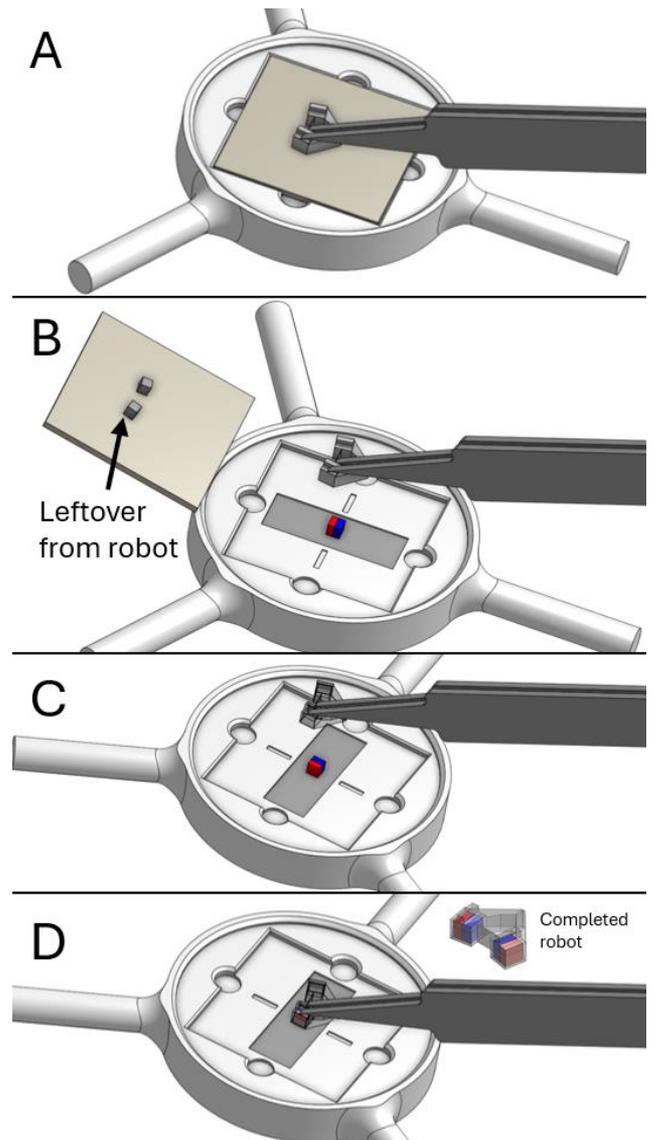

Fig 3: Assembly steps (not to scale). (A). Chassis removed from glass slide. (B). Slide removed from stage, replaced with magnet. (C). Stage rotated to orient magnet. (D). Magnet installed into chassis.

### III. ASSEMBLY PROCESS

The ability to orient a magnetic field and freely pick and place objects makes this system highly versatile. Although it primarily functions as an assembly system for 3D printed robots and block magnets, it is also capable of other pick and place operations. The efficacy of the system was evaluated for three unique operations, all involving objects printed with two photon polymerization on a glass slide (NanoScribe). The first was a simple pick and place task of a 30 µm sphere, and the second and third operations were microassembly tasks. The first microassembly task is shown in Fig 2. It requires a 100 µm cubic permanent magnet to be set in a specific orientation into the head of a helical swimming microrobot [10]. First, the slide is placed onto the rotating stage (Fig 2(A)). The chassis is removed using the tweezers, the slide is set aside, and a micromagnet is placed onto the magnet in the stage (Fig 2(B)). The stage is rotated to align the magnet such that the embedded cubic magnet's poling direction is perpendicular to the helical axis of the microrobot (Fig 2(C)). The chassis is then placed onto the micromagnet (Fig 2(D)). Once the magnet is in place, the assembled robot is lifted above the stage, and cured with UV light to secure the magnet in place.

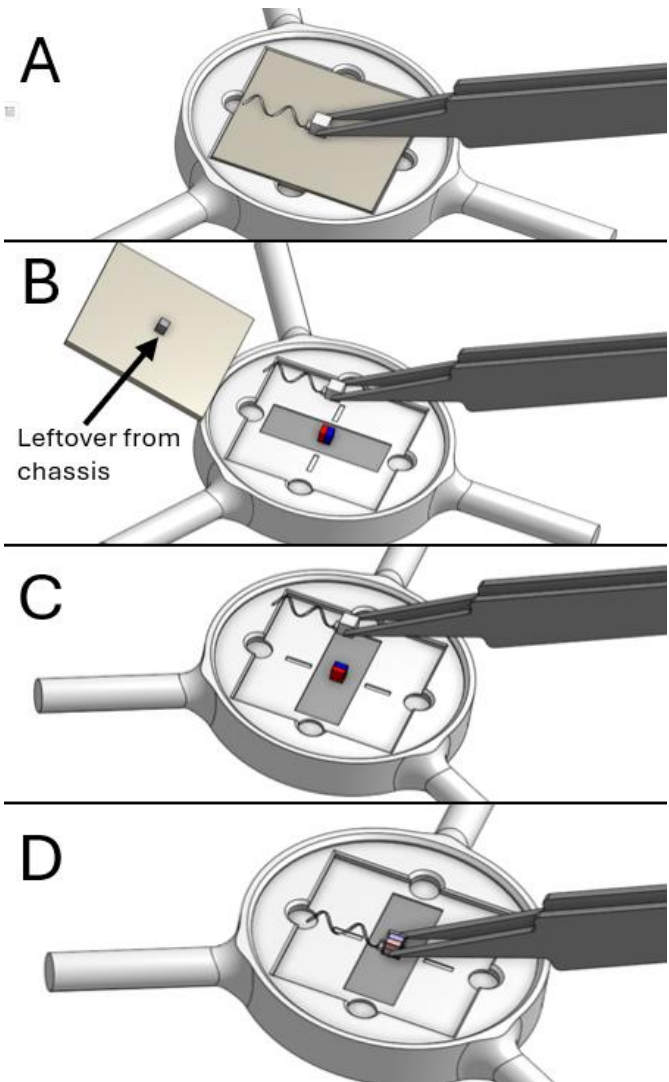

Fig 2: Assembly steps (not to scale). (A). Chassis removed from glass slide. (B). Slide removed from stage, replaced with magnet. (C). Stage rotated to orient magnet. (D). Magnet installed into chassis.

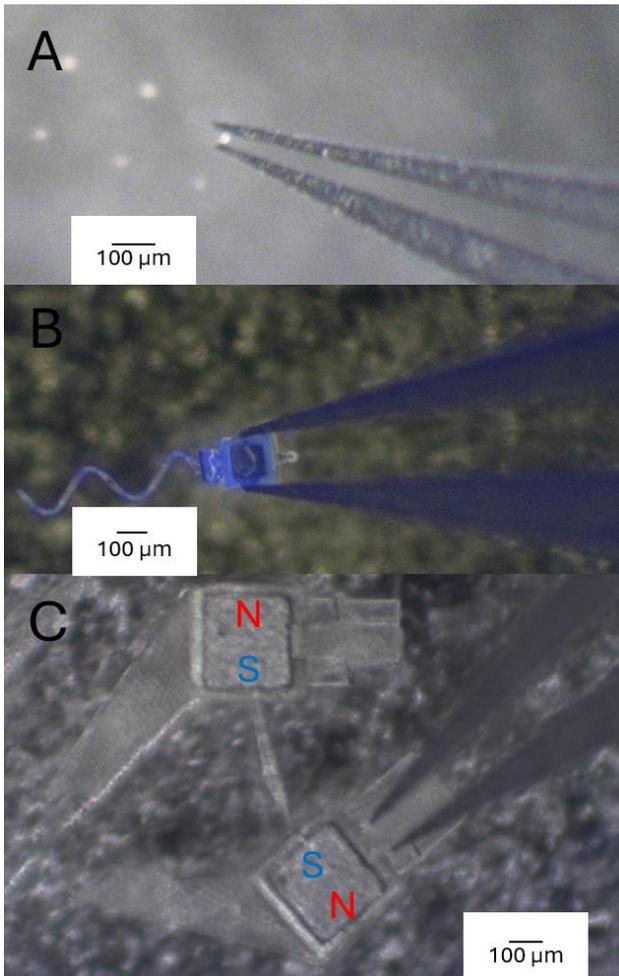

Fig 4: (A). Tweezers grasping and moving a 30 μm sphere. (B). Placement of swimming robot onto a magnet, cured with UV to secure. (C). Fully assembled microgripper robot with magnets oriented in different directions.

The second assembly was for the microgripper robots from [8]. The design requires the installation of two embedded 250 μm cubic magnets in the same plane with opposite polarities. The process is shown schematically in Fig 3. The microrobot chassis is pulled from a slide with the tweezers, using fins integrated into the body for use with this system (Fig 3(A)). The slide is set aside, and a micromagnet is placed onto the magnet in the stage (Fig 3(B)). The stage is rotated to the correct orientation (Fig 3(C)), and the chassis is placed onto the micromagnet (Fig 3(D)). A second magnet is installed into the chassis in the same manner, poled in the opposition direction (Fig 3(D)). Finally, the robot is held above the stage and cured with UV light to secure the magnets in place, completing the assembly.

## IV. RESULTS

Fig 4 shows the results of the evaluation tests. Fig 4(A) shows the pick and place test with a 30 μm sphere. This was the smallest object the system was able to consistently manipulate. The assembled helical swimming microrobot is shown in Fig 4(B). Finally, in Fig 4(C) is an assembled microgripper robot, with two magnets oriented in different directions [4]. These results demonstrate the system's precision and its utility in assembly of magnetic microrobots.

## V. CONCLUSIONS

The microrobotics field continues to yield designs with greater functionality, and magnetic field control has proven to be a promising means by which to operate them. Robots capable of such complex functions require equally complex assemblies. This paper presents a system that fulfills this need and can effectively assemble many of these devices. This development allows for fabrication without the difficulty of acquiring intricate custom parts. This enhances the ability of researchers to test new designs and removes a large barrier to entry into the microrobotics field, which will foster further advancement and innovation.

## ACKNOWLEDGMENT

We would like to thank the helpful faculty and staff at the Birck Nanotechnology Center, and the Purdue College of Engineering, and Mrs. Marilynn Dammon, who has granted the Purdue Bottomley Scholarship on behalf of this research.